\documentclass[sigconf]{acmart}
\usepackage{amsthm}

\AtBeginDocument{%
  \providecommand\BibTeX{{%
    \normalfont B\kern-0.5em{\scshape i\kern-0.25em b}\kern-0.8em\TeX}}}

\newcommand{\cmt}[1]{\ignorespaces}
\newtheorem{proposition}{Proposition}

\copyrightyear{2024}
\acmYear{2024}
\setcopyright{acmlicensed}
\acmConference[KDD '24]{Proceedings of the 30th ACM SIGKDD Conference on Knowledge Discovery and Data Mining}{August 25--29, 2024}{Barcelona, Spain}
\acmBooktitle{Proceedings of the 30th ACM SIGKDD Conference on Knowledge Discovery and Data Mining (KDD '24), August 25--29, 2024, Barcelona, Spain}
\acmDOI{10.1145/3637528.3671858}
\acmISBN{979-8-4007-0490-1/24/08}
\begin{document}

\title{A Learned Generalized Geodesic Distance Function-Based Approach for Node Feature Augmentation on Graphs}

%
\author{Amitoz Azad}
\affiliation{%
  \institution{Singapore Management University}
  \country{Singapore}}
\email{amitoz.sudo@gmail.com}

\author{Yuan Fang}
\affiliation{%
  \institution{Singapore Management University}
  \country{Singapore}}
\email{yfang@smu.edu.sg}

%
%
%
%

\renewcommand{\eqref}[1]{Eq.~(\ref{#1})}

\begin{abstract}
Geodesic distances on manifolds have numerous applications in image processing, computer graphics and computer vision.
In this work, we introduce an approach called `LGGD' (\emph{Learned Generalized Geodesic Distances}).
This method involves generating node features by learning a generalized geodesic distance function through a training pipeline that incorporates training data, graph topology and the node content features. The strength of this method lies in the proven robustness of the generalized geodesic distances to noise and outliers. Our contributions encompass improved performance in node classification tasks, competitive results with state-of-the-art methods on real-world graph datasets, the demonstration of the learnability of parameters within the generalized geodesic equation on graph, and dynamic inclusion of new labels.
\end{abstract}


\begin{teaserfigure}
  \centering
  \includegraphics[width=0.91\textwidth]{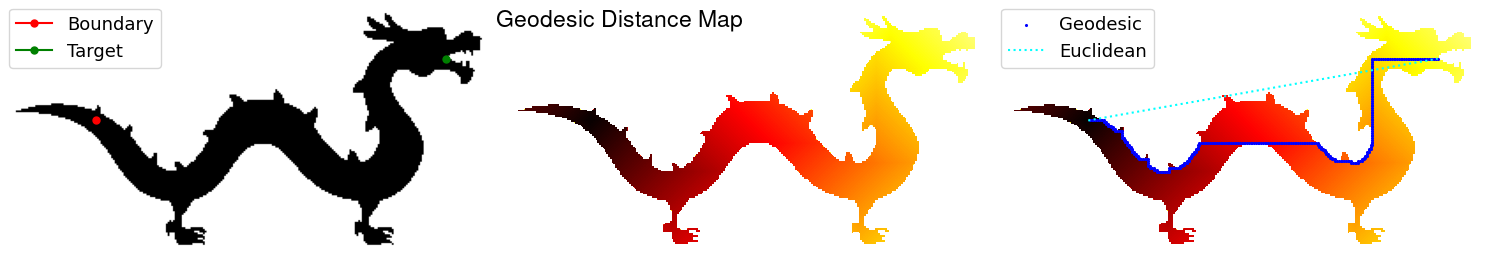}
  \caption{\small (\emph{Left}) A projection of a black dragon onto a 2D grid graph. The aim is to find a geodesic (\emph{shortest-path}) from the source node (boundary) to the target node on the dragon's projection on the grid graph. (\emph{Center}) The geodesic distance map from the source node. (\emph{Right}) The geodesic distance map is used to find the actual geodesic.}
  \label{fig:teaser}
\end{teaserfigure}


\maketitle

\section{Introduction}
In recent times, there has been a growing interest in data augmentation techniques for graphs~\cite{zhao2022graph}. The primary motivation behind augmenting graphs is to improve model performance by enhancing the quality of the graph data through some form of denoising. Real-world graphs, which depict the underlying relationships between nodes, often suffer from noise due to various factors such as fake connections~\cite{hooi2016fraudar}, arbitrary edge thresholds~\cite{tang2018atomistic}, limited or partial observations~\cite{chierichetti2015efficient}, adversarial attacks~\cite{kumar2018false}, and more. These factors collectively render the graphs suboptimal for graph learning tasks. To address these issues, researchers have been exploring graph structural and node feature augmentation techniques that aim to generate improved graphs~\cite{zhao2022graph}.
\cmt{~\cite{gasteiger2019diffusion,zhao2021data,wang2021mixup,verma2021graphmix,rong2019dropedge}.
}

Geodesic distances (Figure ~\ref{fig:teaser}) has found numerous applications in computer vision, ranging from calculating shortest-path distances on discrete surfaces~\cite{kimmel1998computing}, to shape-from-shading~\cite{rouy1992viscosity}, median axis or skeleton extraction~\cite{siddiqi1999hamilton}, graph classification~\cite{borgwardt2005shortest}, statistical data depth~\cite{molina2022tukey}, noise removal, and segmentation~\cite{malladi1996unified}.


Recently the authors in ~\cite{calder2022hamilton} studied a generalized geodesic distance function equation on graphs~\eqref{eq:p-eiko}, which they referred to as the graph $p$-eikonal equation, earlier proposed in~\cite{desquesnes2013eikonal,desquesnes2017nonmonotonic}.
The authors provided both theoretical and experimental evidence to demonstrate that, unlike the geodesic (shortest-path) distance function on graphs (as can be computed with classic Dijkstra algorithm, Sec.~\ref{sec:djk}),  the generalized geodesic distance function is provably more robust (less affected by change) when the graph is subjected to the addition of corrupted edges, especially for $p=1$ in~\eqref{eq:p-eiko}. 



\cmt{~\cite{gasteiger2019diffusion,zhao2021data,wang2021mixup,verma2021graphmix,rong2019dropedge}.
}

\begin{figure*}[t!]
  \centering
  \begin{tabular}{c c c c}
  $n=0$ & $n=10$ & $n=100$ & $n=1000$ \\
    \includegraphics[width=0.2\textwidth]{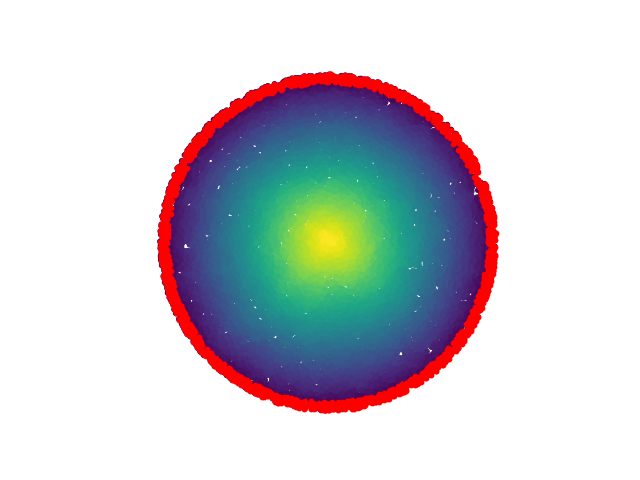} & \includegraphics[width=0.2\textwidth]{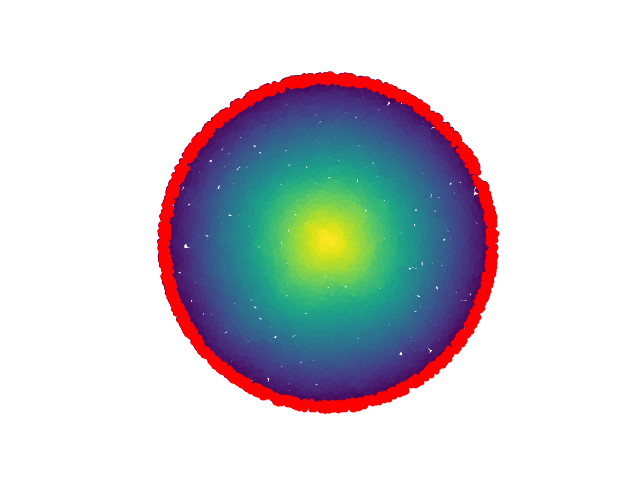} & \includegraphics[width=0.2\textwidth]{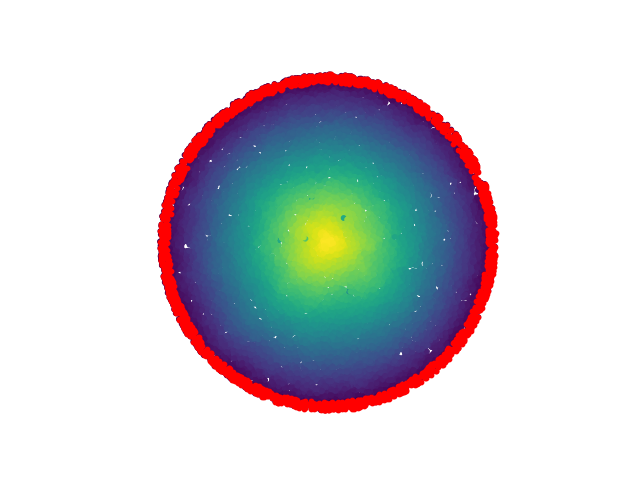} & \includegraphics[width=0.2\textwidth]{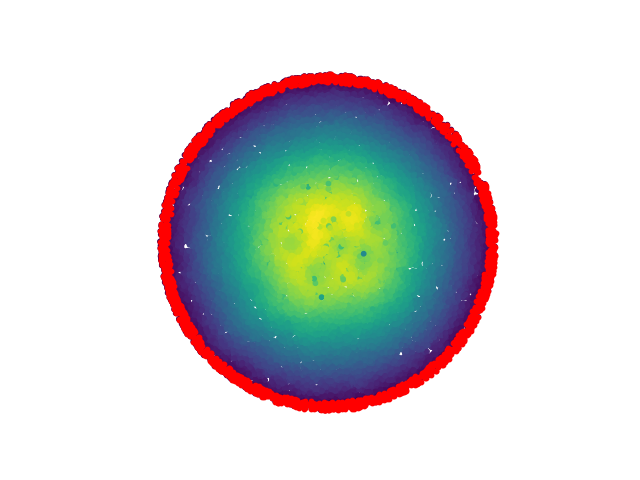} \\
    \multicolumn{4}{c}{\small (a) Robustness of generalized geodesic distance map for $p = 1$ in~\eqref{eq:p-eiko}} \\
    \includegraphics[width=0.2\textwidth]{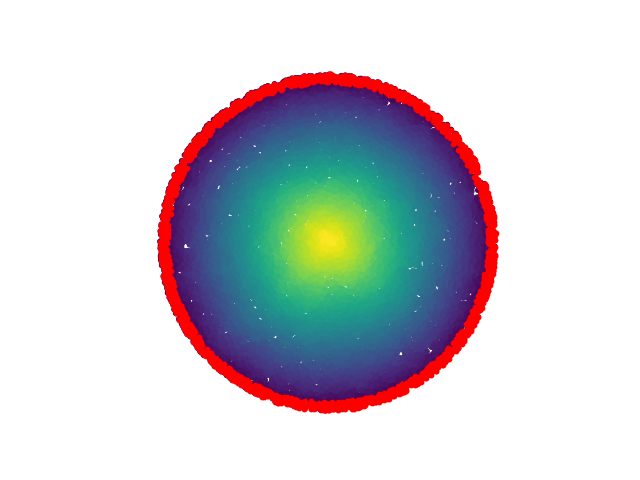} & \includegraphics[width=0.2\textwidth]{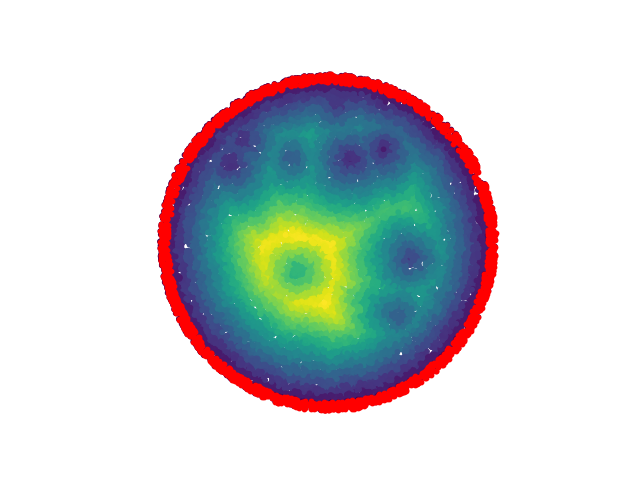} & \includegraphics[width=0.2\textwidth]{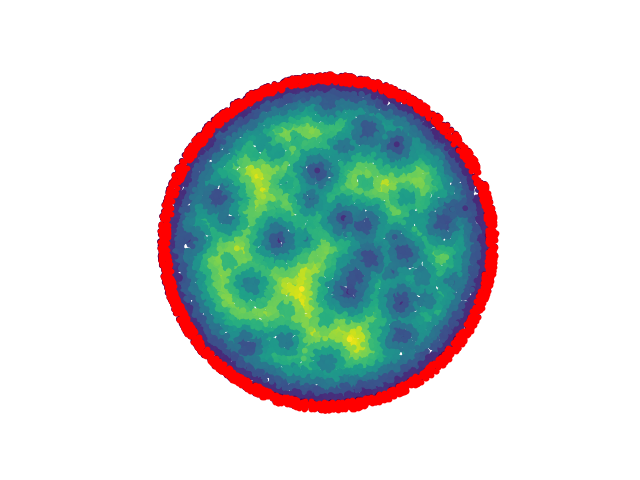} & \includegraphics[width=0.2\textwidth]{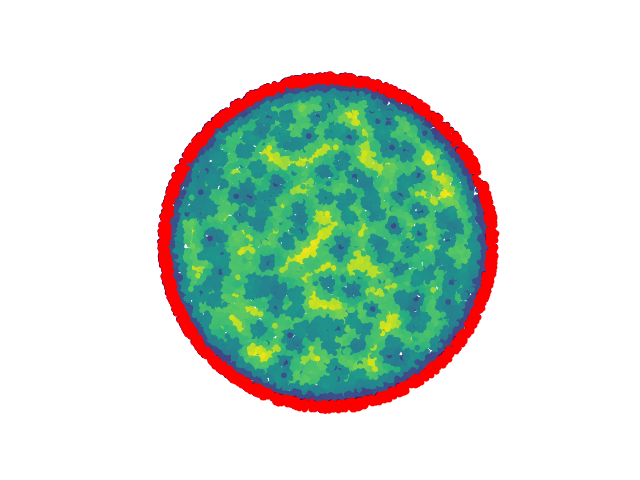} \\
    \multicolumn{4}{c}{\small (b) Robustness of geodesic (shortest-path) distance map ~\eqref{eq:dijk_0}} \\
  \end{tabular}
 
 \caption{\small The $n$ represents the number of random corrupted edges added to a given graph. The graph construction: 20,000 points (nodes) were randomly sampled from a unit ball in ${R}^2$. An $\epsilon$-neighborhood unweighted graph was constructed using these sampled points with $\epsilon=0.05$. All points within $\epsilon$ distance of the boundary of the unit ball are considered boundary nodes. Colors represent the distance from the boundary, with red indicating the boundary where the distance function is zero, and yellow indicating the maximum distance.} 
 
  \label{fg:1}
\end{figure*}

\cmt{The motivation to use $p$-eikonal distances for graph augmentations comes from the demonstrated robustness of $p$-eikonal distances to random edge corruptions (hence noise), as well as the robustness of $p$-eikonal data depth maps in the presence of outliers for density-weighted distances on graphs~\cite{calder2022hamilton}}

\paragraph{Contributions}
Motivated by the proven robustness of the generalized geodesic distance function to edge corruptions (see Figure~\ref{fg:1}) and outliers~\cite{calder2022hamilton}, in this work, we focus on generating node feature vectors using \emph{learned} generalized geodesic distances on a graph for node classification task. This learning of generalized geodesic distances is achieved by formulating the generalized geodesic distance function~\eqref{eq:p-eiko} as a time-dependent problem~\eqref{eq:time}. This time-dependent version allows us not only to solve~\eqref{eq:p-eiko}, but it also enables gradient-based learning of the generalized geodesic distance function using node content features (such as bag-of-words for citation networks).
\cmt{given as input to the backbone model~\mbox{(Table~1 Row~07)}.}

Since the generalized geodesic distance function~\eqref{eq:p-eiko} only considers the graph topology, the generated node features are robust, but they are purely topological, as it does not consider the original node content features. We propose a hybrid model that learns the generalized geodesic distance function using gradient descent, and generates robust node features which not only consider the graph topology but also take into account the original node content features (Figure~\ref{fg:2}). Using these learned generalized geodesic distances at different time values ($t$ in~\eqref{eq:time}) as node features improves the performance of the backbone model, and makes it competitive with state-of-the-art augmentation methods~(Table 1).

We refer to the node feature generated through learning generalized geodesic distances as ``LGGD" (\emph{Learned Generalized Geodesic Distances}) (Table 1). To summarize our key contributions:
\begin{itemize}
    \item  We propose a hybrid model in which the generalized geodesic distances are learned using the training data, graph topology and the node features (Figure~\ref{fg:2}).
    \item The generation of node features based on these learned generalized geodesic distances improves the performance of various backbone models (Figure~\ref{fg:3}, top row) and enables them to compete with SOTA methods (Table~1, Row~09).
    \item The proposed approach allows for the dynamic inclusion of new incoming labels without the need for retraining the backbone GNN {\mbox{(Figure~\ref{fg:3}, bottom row)}}.
    \item We show that gradient based learning of the potential function $\rho(x)$ in generalized geodesic distance function~\eqref{eq:p-eiko} provides a slight boost in the backbone model's performance {\mbox{(Table~1, Row~10)}}.
\end{itemize}

\section{Mathematical Background}
In this section, we review briefly some basic definitions and operators on graphs and provide the necessary background to understand the generalized geodesic distance function on graphs. For a more comprehensive mathematical background, refer to prior works~\cite{ta2008partial,ta2009adaptation}.

\subsection{Notation}
A weighted graph, denoted as \(G = (V, E, w)\), is defined by a finite set of nodes in \(V\) and a finite set of edges in \(E\), where each edge \((i, j)\) connects nodes \(i\) and \(j\). The weights of the graph are determined by a weight function \(w: V \times V \rightarrow [0,1]\), and the set of edges is determined by the non-zero weights: \(E = \{(i, j) | w(i, j) \neq 0\}\). If there is no edge between $i$ and $j$, then $w(i,j)=0$. We represent the set of nodes neighboring node \(i\) as \(N(i)\), where \(j \in N(i)\) signifies that node \(j\) is in the neighborhood of node \(i\), i.e., \(N(i) = \{j \in V | (i, j) \in E\}\).
In this paper, we consider symmetric graphs, meaning that \(w(i, j) = w(j, i)\), and the presence of an edge \((i, j)\) is equivalent to the presence of its reverse \((j, i)\). The degree of a node \(i\), denoted as \(\delta(i)\), is computed as the sum of weights for all nodes in its neighborhood: \(\delta(i) = \sum_{j \in N(i)} w(i, j)\).

Let \(H(V)\) be a Hilbert space comprised of real-valued functions defined on the graph's nodes. A function \(f: V \rightarrow {R}\) of \(H(V)\) characterizes a signal associated with each node, assigning a real value \(f(i)\) to every node \(i\) in \(V\).
Similarly, let \(H(E)\) denote a Hilbert space encompassing real-valued functions defined on the edges of the graph. 

\subsection{Gradient Operators}

The graph difference operator  $d_{w} : H(V) \rightarrow H(E)$ is defined as:
\begin{equation}
  (d_{w}f)(i,j) = \sqrt{w(i,j)}(f(j) - f(i))
\end{equation}
Using this, one can define the graph gradient vector of a function $f \in H(V)$, at a vertex $i \in V$ as:
\begin{equation}
  (\nabla_w f)(i) = [(d_{w}f)(i,j): \forall j \in V]^{T} 
\end{equation}
The $L_p$ norm of the graph gradient is defined as: 
\begin{equation}
\|(\nabla_{w} f)(i)\|_{p} = \big[\sum_{j\in V} |(d_w f)(i,j)|^{p}\big]^{\tfrac{1}{p}}
\end{equation}
Based on the graph difference operator, one can define the directional graph difference operator as follows:   
\begin{equation}
\label{eq:grad}
\begin{split}
  (d^{+}_{w}f)(i,j) &= \sqrt{w(i,j)}(f(j) - f(i))_{+} \\
  (d^{-}_{w}f)(i,j) &= \sqrt{w(i,j)}(f(j) - f(i))_{-}
\end{split}
\end{equation}
Here $(a)_+ = max\{a,0\}$ and $(a)_- = -min\{a,0\}$. 
Following above, one can come up with directional graph gradient vectors as:
\begin{equation}
\begin{split}
  (\nabla^{+}_w f)(i) &= [(d^{+}_{w}f)(i,j): \forall j \in V]^{T}\\
  (\nabla^{-}_w f)(i) &= [(d^{-}_{w}f)(i,j): \forall j \in V]^{T}
\end{split}
\end{equation}
One can then define the $L_p$ norm of these directional graph gradients vectors as:
\begin{equation}
\begin{split}
\|(\nabla^{+}_{w} f)(i)\|_{p} = \big[\sum_{j\in V} |(d^{+}_w f)(i,j)|^{p}\big]^{\tfrac{1}{p}}\\
\|(\nabla^{-}_{w} f)(i)\|_{p} = \big[\sum_{j\in V} |(d^{-}_w f)(i,j)|^{p}\big]^{\tfrac{1}{p}}
\end{split}
\end{equation}
In this work, we focus exclusively on the negative graph gradient operator $(\nabla^{-}_w f)(i)$ and its associated $L_p$ norm $\|(\nabla^{-}_w f)(i)\|_p$ for $p=1$. This operator is closely linked to the morphological erosion PDE process on graphs~\cite{ta2008partial}, and plays a crucial role in defining generalized geodesic distance function on graphs.

{\small 
\begin{figure*}[t!]
\centering
\includegraphics[width=\textwidth]{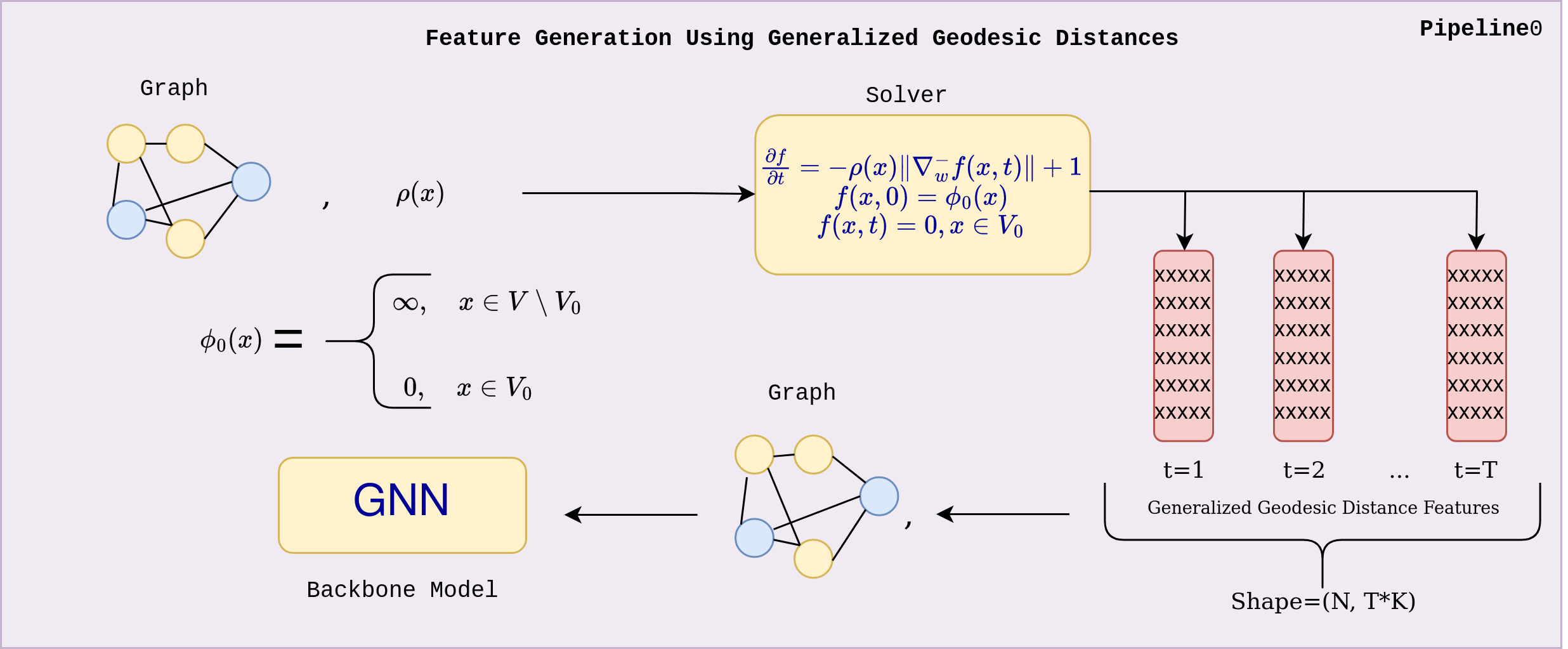}
\caption{Generalized Geodesic Distances as Features}
\label{fg:1.5}
\end{figure*}
}

\subsection{Generalized Geodesic Distance Function}

The generalized geodesic distance function equation on graphs as introduced in~\cite{calder2022hamilton}, can be written in this form (see Appendix A.1):
\begin{equation}
\label{eq:p-eiko}
\begin{split}
\rho(x)\|(\nabla^{-}_w f)(x)\|_p &= 1, \quad x \in V \setminus V_0\\
f(x) &= 0, \quad x \in V_0\\
\end{split}
\end{equation}
Here, $f(x)$ represents the generalized geodesic distance function. $\rho(x)$ is the potential function (Sec. 2.7). $V_0$ represents the boundary nodes (training set) from which the distances have to be calculated (hence $f(x)=0$ at boundary).


One way to solve the above equation is by employing the fast marching~\cite{sethian1999level} or fast iterative~\cite{jeong2008fast} methods. Alternatively, one can solve it numerically by considering a time-dependent version of it:
\begin{equation}
\label{eq:time}
   \begin{split}
       \partial_t f(x,t) &= - \rho(x)\|(\nabla^{-}_w f)(x,t)\|_p + 1, \ x \in V \setminus V_0 \\
       f(x,t) &= 0, \quad \quad \quad \quad \quad \quad \quad \quad \quad \quad \ \ x \in V_0\\
       f(x,0) &= \phi_0(x) \quad \quad \quad \quad \quad \quad \quad \quad \quad \ \!x \in V
   \end{split} 
\end{equation}
At steady-state ($t \rightarrow \infty$), this equation provides the solution to the generalized geodesic distance function~\eqref{eq:p-eiko}.
Note the introduction of an extra variable time $t$ and the initial condition $f(x,0)$. A default choice of initializing is to let the distance be zero on the boundary nodes and infinity (a large positive number) on the rest (\emph{just like the initialization in Dijkstra's algorithm to find the distance map from a source node}).
In this work, we utilize this time-dependent version to generate generalized geodesic distance features for every node for different time ($t$) values. 
As we will see, this formulation provides us with the capability to incorporate the original node content features, and generate \textbf{learned generalized geodesic distances} for different time ($t$) values.
This is achieved by employing backpropagation through an ODE solver~\cite{boltyanskiy1962mathematical}, letting $f(x,0)$ be an MLP (multi-layer perceptron) function of node content features\cmt{($\text{\small MLP}(\text{\footnotesize node feat.})$)}, and then learning the MLP function through gradient descent. 

\subsection{Solving~\eqref{eq:time} with an ODE Solver}
The~\eqref{eq:time} can be solved using an ODE solver like Torchdiffeq~\cite{chen2018neural}. Various numerical schemes, consisting of fixed step or adaptive step sizes, can be employed from Torchdiffeq. Although~\eqref{eq:time} is a PDE on a graph, it can be viewed as a system of coupled ODEs on the graph. This is because the spatial domain is already discretized, and the spatial derivatives at each node can be viewed as finite differences (similar to the Finite Difference Method). 

It must be pointed out that~\eqref{eq:time} is a vector-valued equation. Since the training set (boundary nodes) consists of nodes from $K$ different classes,~\eqref{eq:time} is solved for each class for every node $x$, thus providing $f^k(x,t)$ as the solution. {Here, $f^k(x,t)$ represents the {generalized geodesic distance of node $x$ at time $t$ from the boundary nodes of $k^{th}$ class.}}

\subsection{Learning~\eqref{eq:time} with an ODE Solver}
Torchdiffeq not only allows us to solve a differential equation numerically using various numerical schemes, but it also enables us to learn the parameters of the differential equation through backpropagation  using the adjoint sensitivity method~\cite{boltyanskiy1962mathematical}.

To learn the parameters of the differential equation, first, a segment of the differential equation needs to be converted into a loss function. This loss function is then minimized using a gradient descent based technique via the adjoint sensitivity method. In the case of~\eqref{eq:time}, to learn $\rho(x)$ and $\phi_0(x)$, the boundary condition $f(x,t) = 0$ can be employed to construct a loss $L(f(x,t), 0)$ where $x \in V_0$. 
The approach of converting the boundary condition to a loss function is very similar to the inspiring work done in PINNS~\cite{raissi2019physics}.

\subsection{Connection with Dijkstra} 
\label{sec:djk}
We now explain the connection between~\eqref{eq:p-eiko}, and the celebrated Dijkstra algorithm. The Dijkstra algorithm can be used to find the \emph{geodesic} (shortest-path) distance map on a graph, which then can be further used to find the actual shortest-path between the boundary node and a target node. We call~\eqref{eq:p-eiko} as \emph{generalized geodesic} distance function, because the \emph{geodesic} (shortest-path) distance function (as can be obtained using Dijkstra) is a special case of~\eqref{eq:p-eiko}, as we will see shortly. 

From dynamic programming perspective, for a unweighted graph, the functional equation for Dijkstra algorithm, satisfies the following \emph{shortest-path} distance function from the boundary set $V_0$:
\begin{equation}
\label{eq:dijk_0}
\begin{split}
   f(i) &= \underset{j \in N(i)}{\text{min}} \{f(j) + 1\}, i \in V \setminus V_0\\
   f(i) &= 0, \quad \quad \quad \quad \quad \quad \  i \in V_0
\end{split}
\end{equation}
This equation can then be solved using direct or successive approximation methods~\cite{sniedovich2006dijkstra}. Often the boundary set consist of a single node. The geodesic map on the dragon's projection on the grid (Figure 1) is calculated from the above equation. Once the solution to the above equation is obtained, it enables the determination of the geodesic (shortest-path) from the boundary node(s) and a target node. 
The following proposition explains why~\eqref{eq:p-eiko} is a generalized geodesic distance function.

\begin{proposition}\label{prop:my_prop}
For an unweighted graph with a constant potential function $\rho(x) = 1$, the~\eqref{eq:p-eiko} with supremum norm (\emph{i.e.} $p=\infty$) yields geodesic (shortest-path) distance function of~\eqref{eq:dijk_0}.
\end{proposition}

Refer Appendix A.2 for the proof. The above proposition clearly implies that the space of distance function in ~\eqref{eq:p-eiko} is much larger space which encompasses the geodesic (shortest-path) distance function~\eqref{eq:dijk_0} as a special case. So in that sense, the former represents a more generalized geodesic distance function on graphs.




\subsection{Choosing Potential Function $\rho(x)$}
\label{sec:rho}
The potential function $\rho(x)$, often plays a crucial role in the generalized geodesic distance function of a graph. For instance, in tasks related to image processing, such as segmentation, it is often contingent on the image gradient at a pixel. This dependency allows distances to be shorter in the smooth regions of an image and longer in the non-smooth regions.
In this work, drawing inspiration from~\cite{calder2022hamilton}, we opt to associate the potential function with the local density at a node. By making the potential function dependent on local density, generalized geodesic distances are shortened in denser regions and lengthened in sparser areas. This brings generalized geodesic distances of nodes within a cluster closer together while pushing generalized geodesic distances of nodes in different clusters further apart. We take the node degree, $\delta(x)$, as a measure of density at a node $x$. And set $\rho(x) = \delta(x)^{\alpha}$, where $\alpha$ is a hyperparameter searched within the range of -1 to 0. Later in Sec.~\ref{sec:additional}, we will see that how gradient based learning of this function results in slight boost in the performance.

\cmt{
The speed function rho(x) often plays a crucial role in $p$-eikonal distance map over the graph. For example in image processing related task (like segmentation) it is often dependent on the image graident at a pixel, allowing distances to be shorter in the smooth region of image, and longer at the non-smooth region.
In this work, taking inspiration from [cite], we rather choose it to be associated with local density at a node $i$. 
Making speed function dependent on local density allows the distances 
to be shorter in denser region and longer with sparse region.
This makes the distances of the nodes within cluster closer together,
while driving distances of nodes in different clusters further apart.
We consider the node degree $\delta(i)$ to be measure of density at a node.
And let $\rho(i) = \delta(i)^{\alpha}$, where alpha is a hyperparameter tunned between -1, and 0.
}

\section{Proposed Approaches}
In this section we describe our proposed approaches of generating node features using generalized geodesic distance function without and with gradient based learning.

%
{\small 
\begin{figure*}[t!]
\centering
\includegraphics[width=\textwidth]{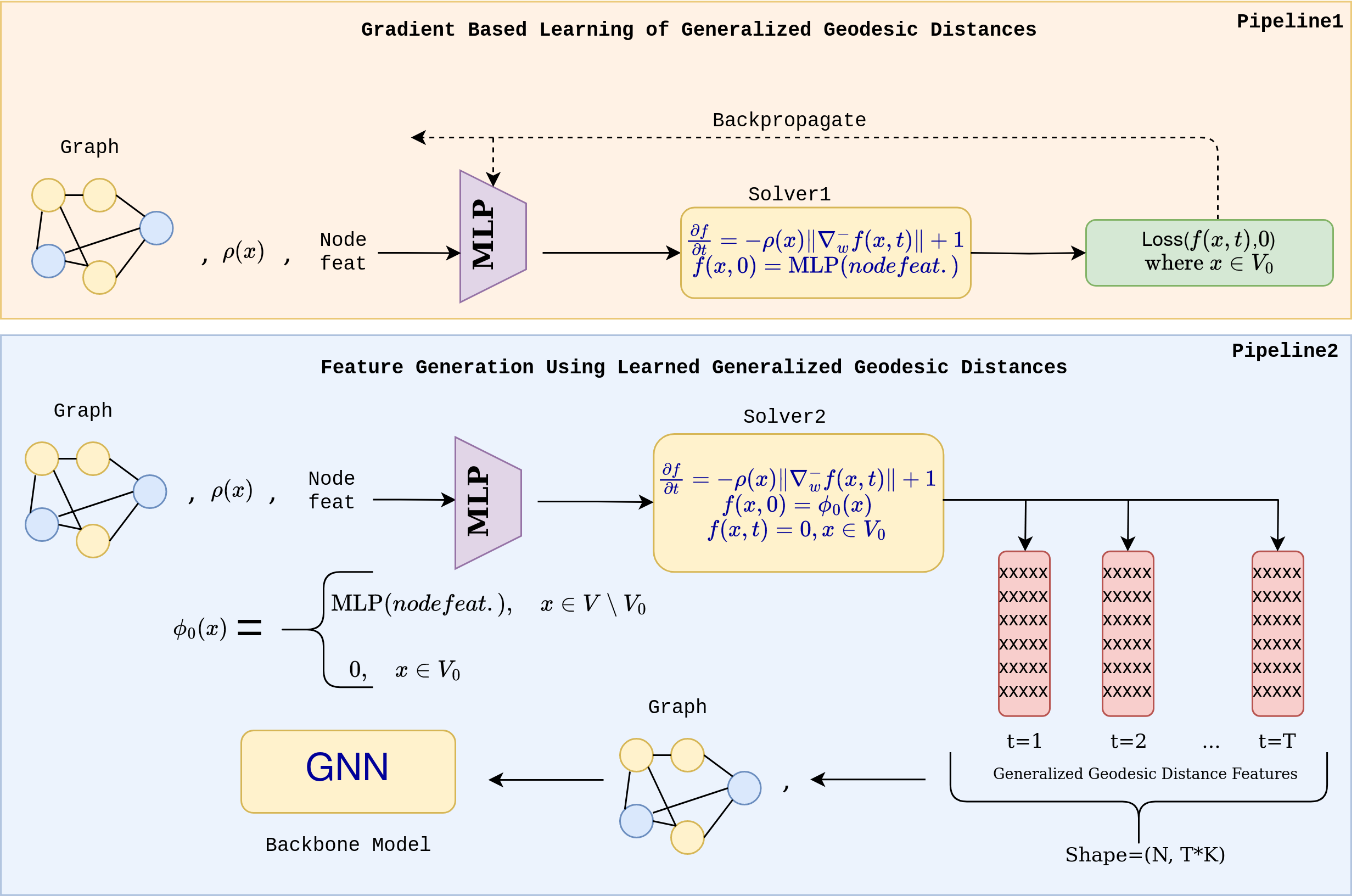}
\caption{Learned Generalized Geodesic Distances as Features.}
\label{fg:2}
\end{figure*}
}

\subsection{Generalized Geodesic Distances as Features}
\label{sec:ed}
This subsection describes the approach to generate generalized geodesic distance as node features with no gradient based learning. To use generalized geodesic distances as node features, we initially start by generating features using~\eqref{eq:time}, where we consider the training set as the boundary nodes and use the default initial condition in~\eqref{eq:time} (Figure~\ref{fg:1.5}). So for the initial condition ($f(x,0)=\phi_0(x), x \in V$) we have:
\begin{equation}
    \phi_0(x) = 
   \begin{cases}
      \infty, \quad \quad \quad \quad \; x \in V \setminus V_0  \\
      0, \quad \quad \quad \quad \ \  x \in V_0
   \end{cases}
\end{equation}
For every node $x$,~\eqref{eq:time} is solved for $K$ different classes in the boundary nodes (training set) for $T$ different time ($t$) values. The final feature at each node $x$ is a set $\{f^k(x,t)\}$, where $k \in \{1,2,..K\}$ and $t \in \{1,2,..T\}$. Here $f^k(x,t)$ represents the generalized geodesic distance of node $x$ at time $t$ from the boundary nodes of $k^{th}$ class. These generalized geodesic distance features are assigned to the nodes (replacing the original node content features) and provided as input to a backbone GNN, along with the graph structure, for the node classification task (Figure~\ref{fg:1.5}). 
We refer to the distance features generated using this approach as \emph{GGD}.

\subsection{Learned Generalized Geodesic Distances as Features}
\label{sec:led}
The previous generalized geodesic distances, treated as node features, are pure topological features as they do not consider the original node content features. This subsection describes the approach to generate the learned generalized geodesic distances. We refer to them as learned generalized geodesic distances (\emph{LGGD}) because they are generated after gradient-based learning of the parameters of~\eqref{eq:time}, as we will see shortly. Figure~\ref{fg:2} depicts the proposed architecture for generating learned generalized geodesic distance features, which also takes the original node content features into account. This architecture can be viewed as a mechanism for converting the original node content features into learned generalized geodesic distance features.

It is essentially a two-step approach involving Pipeline1 and Pipeline2 (Figure~\ref{fg:2}).  Pipeline1 is different from Pipeline0 of the previous case (Figure~\ref{fg:1.5}), as now we have converted the boundary condition into a loss function and added an MLP function as the initial condition (Figure~\ref{fg:2}) in Solver1, which takes into account the original node content features. Unlike Pipeline0, Pipeline1 is not used to generate the distance features; rather, Pipeline1 is tasked with learning the weights of the MLP function and optionally learning the parameters ($\rho(x)$) in Solver1. Node features are input into the MLP, and the output serves as the initial distances $f(x,0)$, provided to Solver1.

The loss function $L(f(x,t), 0)$ within Pipeline1 plays a crucial role in facilitating the learning process. Specifically, it enforces that the self-distances of all nodes on the boundary from the boundary (the training set) should remain zero, as required by the boundary condition in~\eqref{eq:time}. It's worth noting that the boundary condition should not be directly incorporated into Solver1, as doing so would result in the loss remaining perpetually at zero, hindering the learning process (loss minimization using gradient descent).

Pipeline2 serves as the feature-generating pipeline. It is similar to Pipeline0 (no learning case, Figure~\ref{fg:1.5}), as both of them function as feature-generating pipelines. The difference between Pipeline2 and Pipeline0 is that the former uses the learned parameters MLP(\emph{node feat}) and $\rho(x)$ from Pipeline1 to generate the features, whereas the latter uses the default initialization. Observe how the learned MLP function from Pipeline1 is used to construct the initial condition $f(x,0)=\phi_0(x)$ of Solver2:
\begin{equation}
    \phi_0(x) = 
   \begin{cases}
      \text{MLP}(nodefeat.), \quad x \in V \setminus V_0  \\
      0, \quad \quad \quad \quad \quad \quad \quad \  x \in V_0
   \end{cases}
\end{equation}
This construction ensures that self-distances of the boundary nodes are zero from the very beginning $t=0$.

Pipeline2 can be deployed separately once the MLP function and the parameters $\rho(x)$ are learned and saved from Pipeline1. Pipeline2's purpose is to generate learned generalized geodesic distance features for different time steps, which are then concatenated and provided as input to the backbone model for evaluation on the validation and test set. Note that, unlike Solver1, Solver2 explicitly respects the boundary condition specified in~\eqref{eq:time}.

\subsection{Dynamic Inclusion of New Labels}
\label{sec:dynamic}
In Pipeline2, after training the backbone model, one can dynamically include the new labels by simply updating the boundary condition ($f(x,t) = 0, x \in V_0$) and initial condition ($f(x,0) = \phi_0(x), x \in V$) in \eqref{eq:time}. 
Then, one can use the same learned parameters (MLP function and $\rho(x)$ from Pipeline1) to run Solver2 with updated conditions and generate new features for different time steps. These features can subsequently be used as input for the backbone model that has already been trained.

So let $V_1$ be the set of new incoming labels, the new boundary condition would become:  $ f(x,t) = 0, \ x \in (V_0 \cup V_1) $.
And the new initial condition $f(x,0)$ would be:
\begin{equation}
\label{eq:dyn}
    \phi_0(x) = 
   \begin{cases}
      \text{MLP}(nodefeat.), \quad x \in V \setminus (V_0 \cup V_1)  \\
      0, \quad \quad \quad \quad \quad \quad \quad \  x \in V_0 \cup V_1
   \end{cases}
\end{equation}
These updates to boundary condition and initial condition are just to ensure that self-distances to all the nodes on the new boundary ($V_0 \cup V_1$) remain zero for all instances of time $t$.

\section{Experiments \& Results}

\begin{table*}[t]
\label{tab:main}
  \centering
  \caption{Test accuracy over different datasets. From Row 02 to 10, the backbone model is the same GCN. OOM stands for out-of-memory.}
                           {
{\small  

    \begin{tabular}{l@{\hspace{21pt}}r@{\hspace{21pt}}r@{\hspace{21pt}}r@{\hspace{21pt}}r@{\hspace{21pt}}r}
    \hline
    Model & Cora  & Citeseer  & Pubmed  & Computers & Photo \\ 
    \hline
    01 GCN     & 74.13 $\pm$ 2.08   & 66.08 $\pm$ 2.16   & 79.73 $\pm$ 0.71   & 81.72 $\pm$ 1.78    & 87.57 $\pm$ 1.18   \\
    02 MixUp    & 72.72 $\pm$ 1.78  & 64.14 $\pm$ 1.75 & 80.02 $\pm$ 0.52 & 80.76 $\pm$ 1.40    & 88.67 $\pm$ 0.80    \\ 
    03 DropEdge    & 72.28 $\pm$ 1.39  & 65.73 $\pm$ 1.83   & 81.89 $\pm$ 0.84   &  81.45 $\pm$ 1.02   & 88.29 $\pm$ 1.27    \\
    04 GAug-M    & 72.14 $\pm$ 1.37   & 66.38 $\pm$ 1.29   & 82.18 $\pm$ 1.36 & 84.82 $\pm$ 0.78    & 91.05 $\pm$ 1.21  \\ 
    05 GAug-O  & 71.30 $\pm$ 1.54  & 67.22 $\pm$ 1.06 & $\text{OOM}^{*}$ &  83.03  $\pm$ 0.50 & 90.62 $\pm$ 0.30 \\
    06 GDC (heat)    & 77.52 $\pm$ 1.74   & 65.38 $\pm$ 1.36  & 82.16 $\pm$ 0.93  & 80.18 $\pm$ 1.31   & 88.12 $\pm$ 2.21  \\ 
    07 GDC (ppr)    & 78.13 $\pm$ 2.13   & 66.33 $\pm$ 1.84   & 80.86 $\pm$ 0.78   & 82.88 $\pm$ 1.14    & 89.07 $\pm$ 2.19\\ 
    08 GGD & {69.95 $\pm$ 2.51} & {43.21 $\pm$ 2.44}  & {76.49 $\pm$ 0.87} & {78.89 $\pm$ 1.61} & {85.69 $\pm$ 0.92}\\ 
    09 LGGD & {80.18 $\pm$ 1.53} & {67.23 $\pm$ 1.79}  & {83.24 $\pm$ 1.79} & {85.23 $\pm$ 2.18} & {92.02 $\pm$ 2.33}\\ 
    10 LGGD w. $\rho(x)$ & \textbf{81.56 $\pm$ 2.29} & \textbf{68.63 $\pm$ 1.70}  & {83.36 $\pm$ 1.88} & \textbf{85.49 $\pm$ 1.09} & \textbf{92.39 $\pm$ 2.11}\\ 
    11 GPR-GNN    & 79.45 $\pm$ 1.66  & 67.18 $\pm$ 1.84  & \textbf{84.11 $\pm$ 0.38}  & 82.80 $\pm$ 2.01  & 91.48 $\pm$ 1.59 \\
    12 GOAL    & 76.07 $\pm$ 1.56  & 66.57 $\pm$ 1.26  & 81.83 $\pm$ 1.28  & 83.43 $\pm$ 1.04 & 91.65 $\pm$ 0.69 \\ 
    \hline
  \end{tabular}
  }  
  }
\end{table*}

\begin{figure*}[t!]
  \centering
  \begin{tabular}{c c c}
    \includegraphics[width=0.27\textwidth]{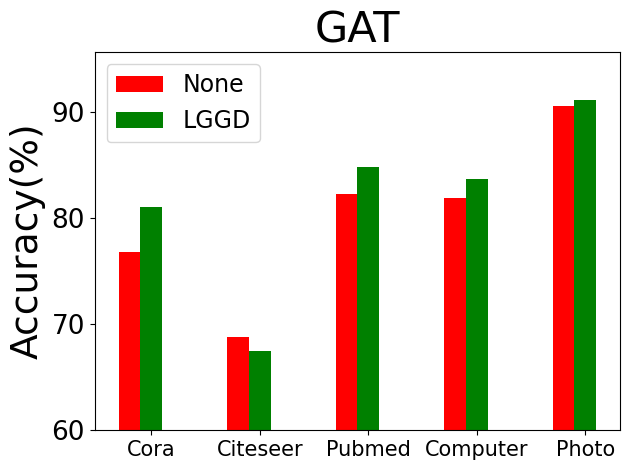} &
    \includegraphics[width=0.27\textwidth]{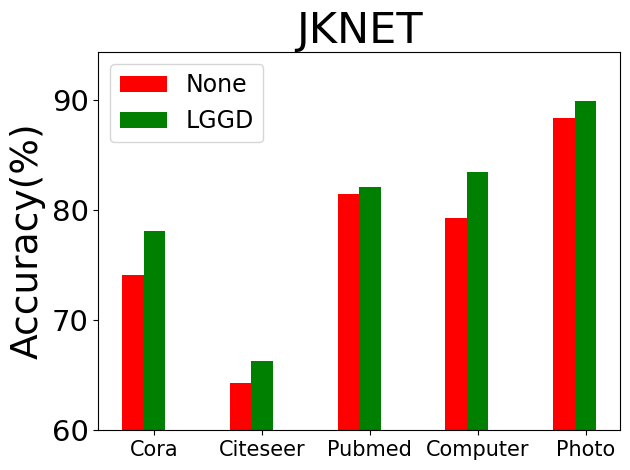} &
    \includegraphics[width=0.27\textwidth]{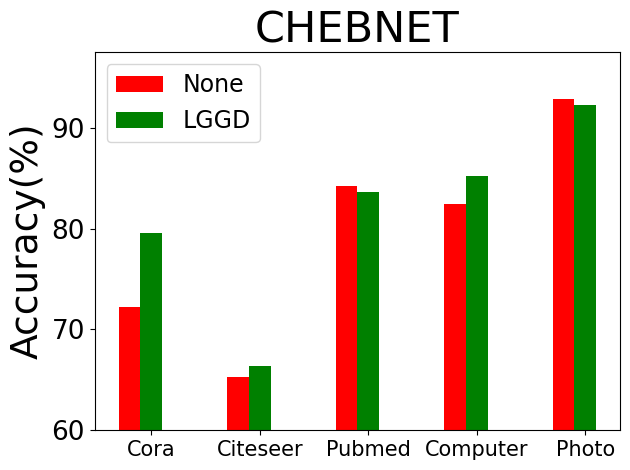} \\
    \includegraphics[width=0.27\textwidth]{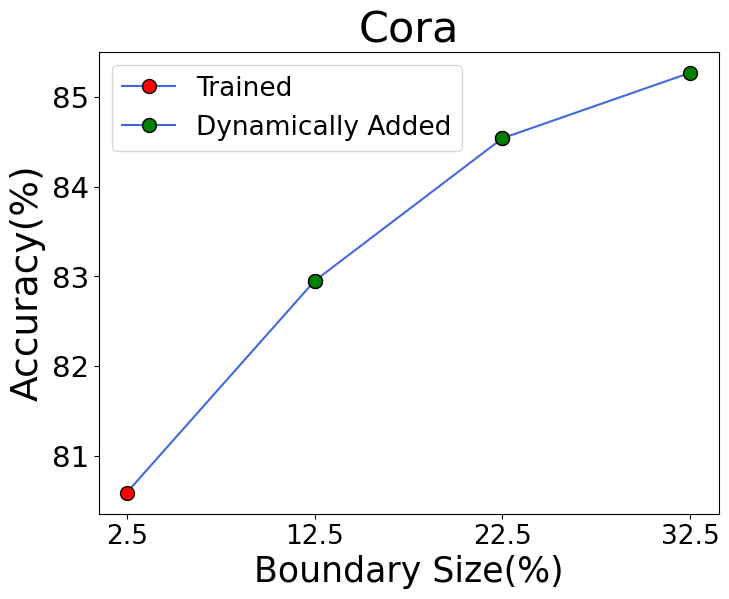} & \includegraphics[width=0.27\textwidth]{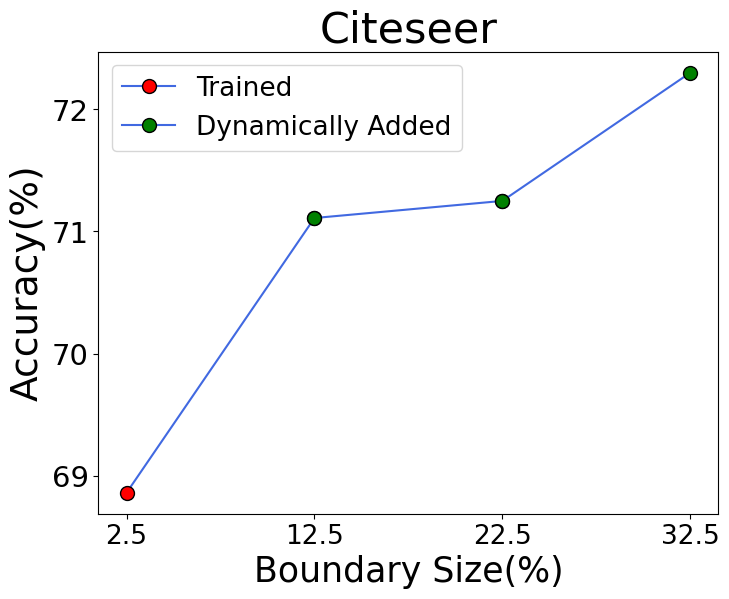} & \includegraphics[width=0.28\textwidth]{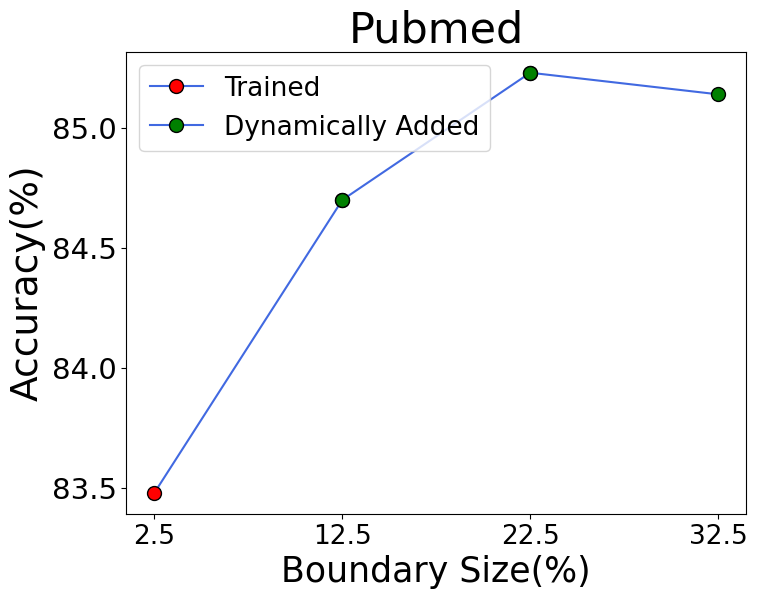} \\
  \end{tabular}
  \caption{\small The top row shows the performance of LGGD (\emph{Learned Generalized Geodesic Distances}) features across different datasets for various backbone models. The bottom row demonstrates the ability to incorporate new incoming labels without retraining the backbone model (see Sec.~\ref{sec:dynamic}), as illustrated for three datasets. A green dot represents the results obtained after dynamically adding 10\% new labels.}
\label{fg:3}
\end{figure*}

In this section, we walk through the research questions pertaining to our proposed models, detail the experiments conducted to answer them, and analyze the results.
\cmt{Unless stated otherwise, whenever we use the term `geodesic distance features,' we are referring to the `generalized geodesic distance features.'}
For all the experiments, we kept $p=1$ in~\eqref{eq:time} as it has been shown to yield the most robust generalized geodesic distances~\cite{calder2022hamilton}.

\paragraph{Software}
We employed the PyTorch framework~\cite{paszke2019pytorch} for our work. To execute the time-dependent generalized geodesic distance~\eqref{eq:time}, we harnessed the combined power of TorchGeometric~\cite{fey2019fast}  along with TorchDiffeq~\cite{chen2018neural}. TorchDiffeq, a well-regarded GPU accelerated ODE solver implemented in PyTorch, offers the capability to perform backpropagation through ODEs using the adjoint sensitivity method~\cite{boltyanskiy1962mathematical}, and it offers a variety of numerical schemes. Throughout all of our experiments, we consistently utilized the Runge-Kutta (RK4) method, adjusting the step size and tolerance values as hyperparameters, which were set using the performance of the backbone model on the validation set.

\paragraph{Datasets}
We used the well-known citation graphs, which have been widely employed for evaluating Graph Neural Networks. These graphs include Cora, Citeseer, and Pubmed~\cite{sen2008collective}, where each node signifies a document, edges represent citation links, and nodes are associated with sparse bag-of-words feature vectors and class labels. In addition to the citation graphs, we incorporated two additional real-world datasets, namely Amazon Photo and Amazon Computer~\cite{shchur2018pitfalls}. In these datasets, nodes represent items, edges signify frequent co-purchases, node features are represented as bag-of-words from product reviews, and the objective is to assign nodes to their respective product categories.

\subsection{Main Results}
\paragraph{RQ1} 
How do the generalized geodesic distance features with no learning~(Sec.~\ref{sec:ed}, Figure~\ref{fg:1.5}) obtained from~\eqref{eq:time} perform on a GCN backbone for node classification?

\paragraph{Evaluation}
For the experiments, we follow a low-resource setting with a train/val set split of 2.5\%/2.5\%, with the rest constituting the test set. We report the average accuracy over 10 random splits. We utilize the performance of the backbone model on the validation set to search the hyperparameters of the ODE solver.
The $\alpha$ in $\rho(x) = \delta(x)^{\alpha}$ (see Sec.~\ref{sec:rho}) was varied in the range 0 to -1 with an interval size of -0.1. In the Runge-Kutta (RK4) numerical scheme the relative tolerance (\emph{rtol}) was varied from 0.001 to 0.05 with an interval size of 0.001. The absolute tolerance (\emph{atol}) in RK4 was always kept as one tenth of the relative tolerance.
The \emph{step\_size} parameter in Runge-Kutta scheme was kept either 0.1 or 1.
The initial distances $f(x,0)$ in~\eqref{eq:time}  are set to be zero for the boundary nodes (training set) and a larger positive value, 1e+6, for the remaining nodes. The features are generated for five different time steps, with $t$ varying from 1 to 5 with an interval of 1. For the backbone GCN~\cite{kipf2016semi}, we employ a hidden layer of size 32, a dropout rate of 0.5, ReLU activation, a fixed learning rate of 0.01, the Adam optimizer, and a weight decay of 1e-6. We train the GCN for 5000 epochs with a patience counter of 100.

\paragraph{Observation}
In Table 1, Row 01 displays the performance of the GCN with original node content features, while Row 08 demonstrates the performance of \emph{Generalized Geodesic Distances} (GGD) as the node features input to the same GCN. It is evident that the GCN using original node features outperforms significantly the use of generalized geodesic distances as input. This observation strongly suggests that the original node content features contain valuable information about the nodes, which the GCN in Row 01 effectively leverages. In contrast, the GGD features represent a purely topological approach and does not take into account any node content features.  Even though the generalized geodesic distances are known to be robust to noise, the original node content features simply outperform GGD features. These findings prompt our next research question.

\paragraph{RQ2}
How do the learned generalized geodesic distance features~(Sec.~\ref{sec:led}, Figure~\ref{fg:2}) perform on a backbone GCN for node classification task? And how do they compare with other graph augmentation methods?

\paragraph{Evaluation}
We use the same splits as in {RQ1}. For the backbone GCN, we retain the settings as before. In the training configuration of Pipeline1, we use 150 epochs and employ the Adam optimizer with a fixed learning rate 0.01 in all of our experiments, along with L2 weight regularization chosen from \{0.0005, 0.001, 0.005, 0.01\}.
The MLP functions used in Pipeline1 had either one or two hidden layers with ReLU activations. The hidden layer size was kept in \{32, 64, 128, 256, 512, 768\}. The dropouts were searched in 0.1 to 0.9 with an interval size 0.1.
The chosen loss function was cross-entropy.  For Pipeline2, we generate features for five different time steps ($t=$1 to 5, with an interval of 1). The hyperparameters for Solver1 and Solver2 were kept same. 
They were searched in the same range as described in the in {RQ1}.
To search for the hyperparameters, we relied on the performance of the backbone model on the validation split. All experiments were conducted using the NVIDIA RTX 3090.

\paragraph{Baselines} 
 We employed various graph augmentation methods for comparision, including MixUp~\cite{wang2021mixup}, DropEdge~\cite{rong2019dropedge}, GDC~\cite{gasteiger2019diffusion}, and GAug~\cite{zhao2021data} (Row 02 to 07 Table 1). One can find more details about these methods in Sec.~\ref{sec:rw}. To learn about the range of their hyperparameter tuning, refer to Appendix A.3. For all these methods, the backbone model remained a simple GCN with the same setting as mentioned before. 
In addition to these models, we also utilized two state-of-the-art models for comparison, namely GPRGNN ~\cite{chien2020adaptive} and GOAL~\cite{zheng2023finding} (Row 11 \& 12, Table 1). 

\cmt{These models do not require the use of a backbone GCN for prediction. We used their available hyperparameters from their respective GitHub sources.}
\cmt{For Mixup, we fine-tuned the $\alpha$ parameter from 1 to 5. The parameter $\lambda$ is sampled from $Beta(\alpha,\alpha)$, which determines the extent of mixing node features. Regarding Dropedge, we adjusted the edge dropout probability from 0 to 0.99. In the case of Gaug-M, we tuned the percentage of the most probable edges to be retained and the percentage of the least probable edges to be dropped, ranging from 0 to 0.9. For Gaug-O, we adjusted the $\alpha$, $\beta$, and $temp$ parameters within the range of [0,1], [0,4], and [0,2] respectively. As for GDC (heat), we varied the parameter $t$ from 1 to 10. In the case of GDC (ppr), we fine-tuned the alpha parameter within the range of 0 to 0.95. For both GDC models, we utilized the \emph{top-k} method to sparsify the influence matrix, selecting from 32, 64, and 128, either along the dimension 0 (row) or dimension 1 (column).}

\paragraph{Observation}
We observe that the proposed model (Figure~\ref{fg:2}) significantly enhances the performance of the generalized geodesic distance features, making it competitive with several other methods (Table 1). Row 08 corresponds to `Generalized Geodesic Distances' (GDD), for which no learning took place. Row 09 corresponds to `Learned Generalized Geodesic Distances' (LGGD), where a significant improvement in the performance of the backbone model is achieved due to the learning factor by incorporating the node content features.

\paragraph{RQ3}
After training the backbone GCN, how does the dynamic inclusion of new labels (Sec.~\ref{sec:dynamic},~\eqref{eq:dyn}) affect performance over test set?

In the bottom row of Figure \ref{fg:3}, one can observe the results of this approach on the three citation networks. We create a split consisting of train/val/nl1/nl2/nl3/test, with percentages of 2.5\%, 2.5\%, 10\%, 10\%, 10\%, and 65\%, respectively. The proposed hybrid model is trained and validated using the 2.5\% splits. After training the backbone model, in the Pipeline2, we dynamically add the new labels (nl1, nl2, nl3) to expand the boundary size ($V_0 = \bigcup_{i=0}^{n} V_i$), and  update the initial condition according to~\eqref{eq:dyn}, and then generate new features using Solver2 and monitor the performance over the test split of the backbone GCN without retraining the backbone GCN. As shown in Figure~\ref{fg:3}, this approach results in a significant increase in performance on the test set without retraining the backbone GCN.

It is important to note that one can always retrain the backbone model with the incoming new labels, possibly achieving even better performances that increase monotonically. However, the purpose of these experiments is to demonstrate faster predictions without retraining. This has practical potential in scenarios where the backbone model is very large and would require a significant amount of time to retrain. In such cases, new predictions can be made in a fraction of a second by simply generating new features with an updated boundary and initial condition \eqref{eq:dyn}, and then providing them as input to the already trained backbone model for inference. 

\subsection{Additional Results}
\label{sec:additional}
\paragraph{RQ4}
How does the optional gradient-based learning of the potential function $\rho(x)$ affect the performance of the generated learned generalized geodesic distances features?

In Row 10 of Table 1, we can observe that this change results in a slight performance increase across all datasets. Making it the the top-performing row across several datasets. It is worth noting that Row 09 and Row 10 share the same hyperparameters, and the slight gains are achieved simply by allowing gradient-based learning of the potential function $\rho(x)$.

\cmt{
How does the gradient based learning of the speed function $\rho(x)$ affect the performance of the learned eikonal distances on the backbone model?
In Row 09 of Table 1, we can observe that this change results in a slight increase in performance across all datasets. Makeing it a top peforming row over several datasets. It is worth noting that Row 08 and Row 09 share the same hyperparameters. The slight gains are achieved simply by allowing the gradient-based learning of $\rho(x)$. 
}

\paragraph{RQ5}
How do the proposed learned generalized geodesic distance features perform for backbone models other than a GCN?

Figure~\ref{fg:3} (top row) showcases the performance of the \emph{LGGD} features on three different backbone models: GAT~\cite{velivckovic2017graph}, CHEBNET~\cite{defferrard2016convolutional}, and JKNET~\cite{xu2018representation}. Mean accuracies for the 10 random splits are presented for the same low-resource split setting. We can observe that the learned generalized geodesic distance-based features lead to performance improvements, sometimes quite significant, on most of the datasets for these models. Refer to Appendix A.4 to know the hyperparams of the backbone models.
\cmt{For GAT, we employed a hidden layer with a size of 32, utilizing input attention heads of size 8 and output attention heads of size 1.
In the case of CHEBNET, we applied a two-step propagation with a hidden layer of size 32. For JKNET, we implemented a GCN model with a hidden layer size of 32. Regarding the layer aggregation component of JKNET, we incorporated a LSTM with 3 layers, each with a size of 16.
For all of these models, the learning rate was set to 0.01, using the Adam optimizer, a weight decay of 1e-6, a dropout rate of 0.5, and a training duration of 5000 epochs with a patience counter set to 100. This configuration was used for  original node content features and for learned generalized geodesic distance distance features.}

While Table 1 aims to demonstrate that our method competes with various state-of-the-art structural and feature augmentation methods (using a common backbone GCN), Figure~\ref{fg:3} (top row) illustrates how our method enhances the performances across different backbone GNNs.
It is essential to note that, for both Table 1 and Figure~\ref{fg:3}, the hyperparameter configuration of the backbone GNN is consistently maintained (with and without augmentation(s)), allowing us to focus solely on studying the effect of augmentation.

\cmt{
How does the proposed LED (Learned Eikonal Distance) features perform for backbone models other than a GCN?

Figure 2 displays the performance of the learned Eikonal distances on three different backbone models: GAT [cite], CHEBNET [cite], and JKNET [cite]. Mean accuracies for the 10 random splits are presented for the same low-resource split setting.
We can observe that the learned Eikonal distance-based features result in performance improvement, sometimes quite significant, on most of the datasets for these models.
}

\section{Related Work}
\label{sec:rw}

The field of graph augmentation is vast and rapidly gaining interest within the graph learning community. Here, we will focus on some popular methods for node-level tasks, which essentially make the graph robust to noise by either changing its topology (structural augmentation) or altering its node features.

GraphMix~\cite{verma2021graphmix} and MixUp~\cite{wang2021mixup} are two popular methods for node feature augmentation in semi-supervised learning. Both GraphMix and MixUp employ training a Graph Neural Network (GNN) by interpolating node features and node targets using a convex combination. Both methods draw the parameter $\lambda$ for the convex combination from a beta distribution. While MixUp involves the mixing of node features and their hidden representations through the message passing within a GNN, GraphMix utilizes a Fully Connected Network (FCN) alongside a GNN, exclusively for feature mixing. The FCN layers and GNN layers share their weights and are jointly trained on a common loss function, combining predictions from the training set FCN layer and GNN layer. Additionally, an unsupervised loss term is incorporated to ensure that the predictions of the GNN on unlabeled nodes match those of the FCN.

DropEdge~\cite{rong2019dropedge}, GAug~\cite{zhao2021data} and GDC~\cite{gasteiger2019diffusion} are three popular structural augmentation methods for node classification.  DropEdge just randomly removes the edges, and redo the normalization on the adjacency matrix before every training epoch. GAug comes in two versions: GAug-M and GAug-O. Both use Graph Autoencoder as the edge prediction module. In GAug-M, an edge prediction module is trained before passing the modified graph to the backbone model. Then, edges with high and low probabilities are added and removed, respectively. In GAug-O, the edge prediction module is trained in combination with the backbone model, using a common loss function that combines node classification loss and edge prediction loss. Training is performed by sparsifying the convex combination of the edge prediction module and the original graph, using differential Bernoulli sampling on this combination. We find it to be slow and memory intensive (Table 1). 
GDC essentially smooths out the neighborhood by acting as a denoising filter, similar to a Gaussian filter for images. It achieves this by first calculating an influence matrix using methods such as page rank or a heat kernel to make the graph fully connected. Then, it sparsifies the influence matrix using either a \emph{top-k} cutoff or an epsilon cutoff to retain only the edges with maximum influence.

\section{Conclusion}
We proposed a hybrid model in which learned generalized geodesic distances were used as node features to improve the performance of various backbone models for the node classification task. The proposed model allows the dynamic inclusion of new incoming labels without retraining the backbone model.

One limitation of our work is that we did not find much success for heterophilous graph datasets. In fact, most of the structural and node feature augmentation methods work only on homophilous graph datasets.
One potential way to overcome this issue is to try negative weights to represent dissimilarity~\cite{ma2016diffusion}. Alternatively, allowing the potential function to take on negative values could be considered. These approaches will be investigated in the future.
\cmt{The runtime complexity of proposed model is dominated by the $O(|E|)(F_f + F_b)$.}

\begin{acks}
This research project is supported by the Ministry of Education, Singapore, under its Academic Research Fund Tier 2 (Proposal ID: T2EP20122-0041). Any opinions, findings and conclusions or recommendations expressed in this material are those of the author(s) and do not reflect the views of the Ministry of Education, Singapore.
\end{acks}

\bibliographystyle{ACM-Reference-Format}
\balance
\bibliography{sample-base}

\appendix

\section{Appendices}
\subsection{Rewriting Graph $p$-Eikonal Equation in~\cite{calder2022hamilton} as Eq.~(\ref{eq:p-eiko})}
Let us recall the generalized geodesic distance function equation: 
\begin{equation*}
\begin{split}
\rho(i)\|(\nabla^{-}_w f)(i)\|_p &= 1, \quad i \in V \setminus V_0\\
f(i) &= 0, \quad i \in V_0\\
\end{split}
\end{equation*}

Using the definitions of $\|(\nabla^{-}_w f)(i)\|_p$ and $(d^{-}_wf)(i,j)$ from Sec 2.2, one obtains the following for the non-boundary nodes ($V \setminus V_0$):
\begin{equation*}
\begin{split}
   \sum_{j\in V} {w(i,j)}^{\tfrac{p}{2}}(f(j) - f(i))^{p}_{-}  &= (\rho(i))^{-p}\\
\end{split}
\end{equation*}
Using $(a)_- = -min\{a,0\} = max\{-a,0\} = (-a)_+$:
\begin{equation*}
\begin{split}
   \sum_{j\in V} {w(i,j)}^{\tfrac{p}{2}}(f(i) - f(j))^{p}_{+}  &= (\rho(i))^{-p}\\
\end{split}
\end{equation*}
By introducing a change of variable $w(i,j)^{\tfrac{p}{2}} = \Tilde{w}(i,j)$ and $\rho(i)^{-p} = \Tilde{\rho}(i)$, one obtains the exact graph $p$-eikonal equation as proposed in~\cite{calder2022hamilton}.

\subsection{Proof of Proposition1}
%
\paragraph{Proof} Let us recall that the supremum norm (\emph{aka} infinity norm) for $n$ dimensional vector $x$ is $\|x\|_{\infty} = max\{|x_1|, |x_2|, ... |x_n|\}$. By using $\|.\|_{\infty}$ norm in generalized geodesic distance function~\eqref{eq:p-eiko} becomes: 
\begin{equation*}
\begin{split}
\underset{j \in V}{\text{max}} \{|(d^{-}_w f)(i,j)|\}  &=  (\rho(i))^{-1}, \quad i \in V \setminus V_0\\
f(i) &= 0, \quad i \in V_0\\
\end{split}
\end{equation*}
For a unweighted graph with potential function $\rho(i) = 1$, one obtains the following for the non boundary nodes $V \setminus V_0$:
\begin{equation*}
\begin{split}
 \underset{j \in N(i)}{\text{max}} \{\text{max}(0, f(i) - f(j))\}  &= 1\\
\end{split}
\end{equation*}
Here we used $(a)_- = (-a)_+$, and $w(i,j)=0$ when $j \notin N(i)$.

One can rewrite the above as:
\begin{equation*}
\begin{split}
\underset{j \in N(i)}{\text{max}} \{\text{max}(-1, f(i) - f(j) - 1)\}  &= 0\\
\end{split}
\end{equation*}
Since the R.H.S. is zero, for any valid solution of the above equation, there must be at least one $j \in N(i)$ for which $(f(i) - f(j) - 1)$ is zero, and this corresponds to the maximum element in the set on L.H.S. Therefore, the above equation can be rewritten as:
\begin{equation*}
\begin{split}
 \underset{j \in N(i)}{\text{max}} \{(f(i) - f(j) - 1)\}  &= 0\\
 f(i) - \underset{j \in N(i)}{\text{min}} \{(f(j) + 1)\} &= 0. \\
\end{split}
\end{equation*}
The above equation corresponds exactly to~\eqref{eq:dijk_0}.

\subsection{Hyperparam Tuning of Baselines}
{For MixUp~\cite{wang2021mixup}, we used random search to uniformly draw $\alpha$ parameter from 1 to 5. The parameter $\lambda$ is sampled from $Beta(\alpha,\alpha)$, which determines the extent of mixing node features. Regarding DropEdge~\cite{rong2019dropedge}, we adjusted the edge dropout probability from 0 to 0.99 with an interval size of 0.01. In the case of GAug-M~\cite{zhao2021data}, we tuned the percentage of the most probable edges to be retained and the percentage of the least probable edges to be dropped, ranging from 0 to 0.9 with an interval size of 0.1. For GAug-O~\cite{zhao2021data}, we used random search to uniformly draw $\alpha$, $\beta$, and $temp$ parameters within the range of [0,1], [0,4], and [0,2] respectively. As for GDC (heat)~\cite{gasteiger2019diffusion}, we varied the parameter $t$ from 1 to 10 with and interval of 0.5. In the case of GDC (ppr)~\cite{gasteiger2019diffusion}, we fine-tuned the alpha parameter within the range of 0 to 0.95 with an interval of 0.05. For both GDC models, we utilized the \emph{top-k} method to sparsify the influence matrix, selecting from 32, 64, and 128, either along the dimension 0 (row) or dimension 1 (column).}
{GPRGNN~\cite{chien2020adaptive} and GOAL~\cite{zheng2023finding} do not require the use of a backbone model for prediction. We used their available hyperparameters from their respective GitHub sources.}

\subsection{Hyperparams of Different Backbones}
For GAT, we employed a hidden layer with a size of 32, utilizing input attention heads of size 8 and output attention heads of size 1.
In the case of CHEBNET, we applied a two-step propagation with a hidden layer of size 32. For JKNET, we implemented a GCN model with a hidden layer size of 32. Regarding the layer aggregation component of JKNET, we incorporated a LSTM with 3 layers, each with a size of 16. For all of these models, the learning rate was set to 0.01, using the Adam optimizer, a weight decay of 1e-6, a dropout rate of 0.5, and a training duration of 5000 epochs with a patience counter set to 100. This configuration was used for  original node content features and for learned generalized geodesic distance distance features.

\subsection{Efficiency}
    \paragraph{Training Time}
    The overall training time depends on that of the backbone in Pipeline2. Regarding backpropagation through the ODE solver (Pipeline1), the training time efficiency (training loss vs time) is known to be a few times lower than an MLP (as shown in the Figure 20 of work done by Dupont \emph{et al.}~\cite{dupont2019augmented}). 
    However, this can be effectively mitigated by  concatenating every feature vector with zeros~\cite{dupont2019augmented}.
    \cmt{However, this can be effectively mitigated by employing an augmentation technique of concatenating every feature vector with zeros~\cite{dupont2019augmented}, which we did not utilize.}
    
    \paragraph{Complexity}  The runtime complexity of the ODE solver is dominated by $O(|E|k)(F_b + F_f)$. Here, $|E|$ represents the number of edges, $k$ represents the number of classes, and $F_b$ and $F_f$ represent the numbers of backward and forward function evaluations, respectively. The complexity of the backbone GNN depends on the specific backbone.

    \paragraph{Inference}  Once the MLP in Pipeline1 is trained, it can quickly produce learned generalized geodesic distance features, taking only a fraction of a second. The primary benefit for the dynamic inclusion part is the ability to make fast predictions without needing to retrain the backbone GNN model. For instance, the prediction times for the dynamically added new labels case (green dots in Figure 5, bottom row) is around 0.1 sec for all three citation graphs, whereas retraining a simple backbone model like a GCN for 1k epochs takes around 7 sec for Pubmed dataset on RTX 3090.
    
    \paragraph{Scalability}  The overall scalability depends on the scalability of the backbone GNN. Concerning the ODE solver's ability to manage large-scale graphs, it is worth noting that it has been successfully utilized in the literature~\cite{chamberlain2021grand, chamberlain2021beltrami} for handling OGB graphs in node classification task. 
    
\end{document}